\documentclass{article}

\usepackage{PRIMEarxiv}

\usepackage[utf8]{inputenc} 
\usepackage[T1]{fontenc}    
\usepackage{hyperref}       
\usepackage{url}            
\usepackage{booktabs}       
\usepackage{amsfonts}       
\usepackage{nicefrac}       
\usepackage{microtype}      
\usepackage{lipsum}
\usepackage{fancyhdr}       
\usepackage{graphicx}       
\graphicspath{{media/}}     

\title{KAN-Mixers: a new deep learning architecture for image classification}

\author{
  Jorge Luiz dos Santos Canuto\\
  Maringá State University \\
  Maringá, Paraná, Brazil\\
  \texttt{pg404818@uem.br} \\
  \And
  Linnyer Beatrys Ruiz Aylon \\
  Maringá State University \\
  Maringá, Paraná, Brazil\\
  \texttt{lbruiz@uem.br} \\
   \AND
  Rodrigo Clemente Thom de Souza \\
  Paraná Federal University \\
  Maringá State University \\
  Maringá, Paraná, Brazil\\
  \texttt{thom@ufpr.br} \\
}

\begin{document}
\maketitle

\begin{abstract}
Due to their effective performance, Convolutional Neural Network (CNN) and Vision Transformer (ViT) architectures have become the standard for solving computer vision tasks. Such architectures require large data sets and rely on convolution and self-attention operations. In 2021, MLP-Mixer emerged, an architecture that relies only on Multilayer Perceptron (MLP) and achieves extremely competitive results when compared to CNNs and ViTs. Despite its good performance in computer vision tasks, the MLP-Mixer architecture may not be suitable for refined feature extraction in images. Recently, the Kolmogorov-Arnold Network (KAN) was proposed as a promising alternative to MLP models. KANs promise to improve accuracy and interpretability when compared to MLPs. Therefore, the present work aims to design a new mixer-based architecture, called KAN-Mixers, using KANs as main layers and evaluate its performance, in terms of several performance metrics, in the image classification task. As main results obtained, the KAN-Mixers model was superior to the MLP, MLP-Mixer and KAN models in the Fashion-MNIST and CIFAR-10 datasets, with 0.9030 and 0.6980 of average accuracy, respectively.
\end{abstract}


\section{Introduction}
Computer vision is a field of artificial intelligence that encompasses methods and techniques that provide machines with the ability to learn from image data. This area of computer science includes software, hardware, and imaging techniques required for such methods \cite{patricio_computer_2018}. In this context, there are several computer vision tasks that can be solved by machines and that find applications in various areas of society, namely: engine fault diagnosis \cite{dahmer_dos_santos_thermographic_2023}, astronomy \cite{costa-duarte_s-plus_2019}, human-computer interface \cite{cizotto_web_2023}, object detection \cite{ahmad_lightweight_2025, peng_improved_2025}, facial recognition \cite{alansari_efficientfacev2s_2025}, among others. In addition, several deep learning models are proposed to solve such tasks.

With their architecture based on convolutional layers, Convolutional Neural Networks (CNNs) \cite{Lecun1998} dominated computer vision tasks for a few years. Recently, Transformer-based architectures, specifically Vision Transformer (ViT) \cite{Dosovitskiy2020} and Swin Transformer \cite{Liu2021}, emerged as an alternative based on self-attention layers, a mechanism that learns relationships between different image patches. Thus, Transformers have demonstrated attractive performance, often outperforming CNNs, especially on large datasets \cite{Tolstikhin2021,Trockman2022a,li_how_2023}.

In 2021, Google proposed MLP-Mixer \cite{Tolstikhin2021}, a more concise visual architecture with higher inference speed than ViT. Despite its simple structure, which relies only on Multilayer Perceptron (MLP), MLP-Mixer achieves extremely competitive results, as demonstrated in Tolstikhin (2021). When pre-trained on large datasets, MLP-Mixer achieves an accuracy/cost ratio almost equal to the state-of-the-art performance claimed by CNNs and Transformers \cite{Zhang2022}. However, although MLP-Mixer presents competitive performance, this architecture may be inadequate for refined feature extraction and requires a large amount of training data \cite{Liu2024}.

Recently, the Kolmogorov-Arnold Network (KAN) has been proposed as a more attractive alternative to MLP models. KAN replaces the fixed activation functions of MLP models with learnable activation functions \cite{Peng2024}. This architectural difference allows the KAN model to learn and also enhance useful features of the data \cite{Liu2024a}. Furthermore, KANs can lead to remarkable accuracy and improved interpretability compared to MLPs \cite{Liu2024a}.Therefore, this work developed a new computer vision architecture based on KAN and the MLP-Mixer architecture, called KAN-Mixers.

\section{Materials and Methods}
\label{sec:Materials and Methods}

\subsection{Fashion-MNIST Dataset}
The Fashion-MNIST dataset consists of 70,000 grayscale images in the 28x28 format. These images represent 10 categories of fashion products, with 7,000 images per category. The training set contains 60,000 images and the test set has 10,000 images \cite{xiao_fashion-mnist_2017}. Fashion-MNIST provides an initial challenge to computer vision models and is often used as a benchmark for machine learning algorithms. Figure 1 shows some images from the Fashion-MNIST set.

\begin{figure}[!ht]
  \centering
  \includegraphics[width=0.45\textwidth]{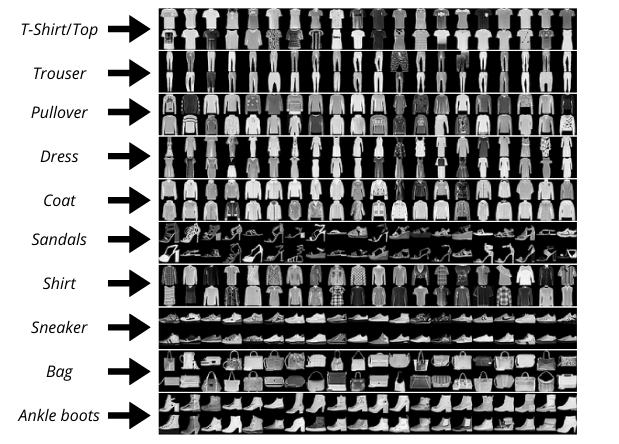}
  \caption{Examples from the Fashion-MNIST dataset.}
  \label{fig:fig1}
\end{figure}

\subsection{CIFAR-10 Dataset}
CIFAR-10 is a widely used benchmark dataset in the computer vision community. This dataset contains 60,000 color images with a resolution of 32x32, divided into 50,000 images for training and 10,000 images for testing. This dataset has 10 classes, with 6,000 images in each class. An example of the images from the CIFAR-10 dataset is available in \cite{Thakkar2018} and can be seen in Figure 2.

\begin{figure}[!ht]
  \centering
  \includegraphics[width=0.70\textwidth]{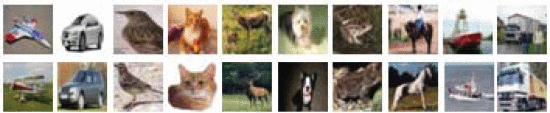}
  \caption{Examples from the CIFAR-10 dataset.}
  \label{fig:fig2}
\end{figure}

\subsection{Description of the KAN-Mixers architecture}
The main deep computer vision architectures consist of layers that mix features in a given spatial location and layers that mix features between different spatial locations, or layers that do both tasks at the same time. The idea behind Mixer-based architectures is to clearly separate operations per location (channel-mixing) and operations between locations (token-mixing) \cite{Tolstikhin2021, Trockman2022a}.

Therefore, the KAN-Mixers architecture performs channel-mixing and token-mixing using only KAN models. The KAN-Mixers architecture consists of a patch-embedding layer, repeated applications of the Mixer Block, an Adaptive Average Pooling layer and, finally, a Classifying Head. The architecture of the KAN-Mixers model can be seen in Figure 3.

\begin{figure}[!ht]
  \centering
  \includegraphics[width=0.90\textwidth]{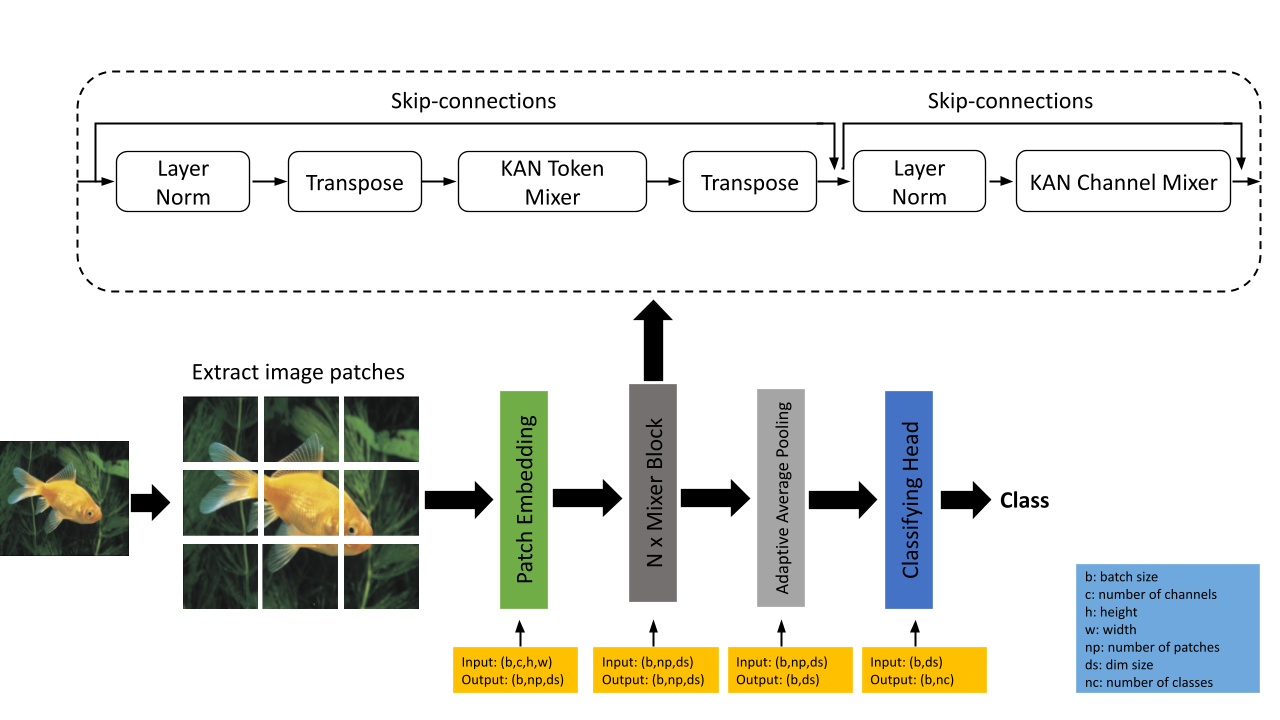}
  \caption{KAN-Mixers Architecture.}
  \label{fig:fig3}
\end{figure}

As seen in Figure 3, patches are first extracted from the image, then the patches pass through an embedding layer, followed by repeated application of the Mixer Block, finishing with an Adaptive Average Pooling layer and the output generated by the Classifying Head. In addition, it can also be seen that the Mixer Block is formed by Layer Normalization, transposition, KAN Token Mixer and KAN Channel Mixer layers. The KAN Token Mixer and KAN Channel Mixer are implemented through KAN and Dropout layers.

The core of the KAN-Mixers model is the Mixer Block which, similarly to the MLP-Mixer model, can be described mathematically as:

\begin{equation}
U = X^{T}_{Columns} + W_{2}(W_{1}\textit{LayerNormalization}(X^{T}_{Columns}))
\end{equation}
\begin{equation}
Y = U^{T}_{rows} + W_{4}(W_{3}\textit{LayerNormalization}(U^{T}_{rows}))
\end{equation}

Where X is the input and $W_{1}, W_{2}, W_{3}$ e $W_{4}$ are KAN layers. Equation 1 shows the KAN layers acting on the transposed input X for token mixing and Equation 2 represents the KAN layers acting on the transposed input U for channel mixing. It is important to highlight that unlike other Mixer architectures, such as MLP-Mixer and ConvMixer, KAN-Mixers does not rely on a fixed activation function, as it is a fully KAN-based architecture. Furthermore, given that KAN layers are slower to train, all KAN layers used in this work are an efficient implementation of the original KAN, available at https://github.com/Blealtan/efficient-kan.

\subsection{Experiment description}
Figure 4 presents the methodology adopted in this work, where it is possible to observe the datasets used, the proposed KAN-Mixers model, the MLP-Mixer, MLP and KAN models, which were compared with the model proposed in this work, and, finally, the result obtained after training the models, in terms of average accuracy.

\begin{figure}[!ht]
  \centering
  \includegraphics[width=0.60\textwidth]{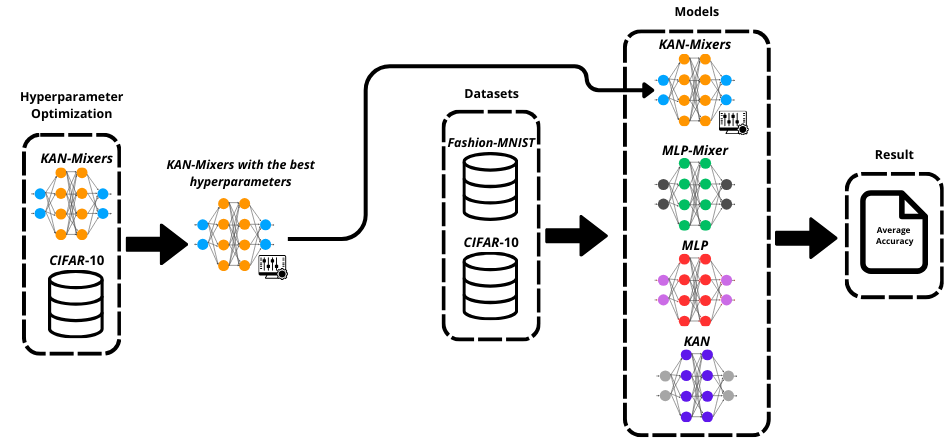}
  \caption{Proposed methodology.}
  \label{fig:fig4}
\end{figure}

Initially, as shown in Figure 4, a hyperparameter optimization step was performed using the KAN-Mixers model and the CIFAR-10 dataset. As a result of this step, a KAN-Mixers model with the best hyperparameters was obtained. Next, the Fashion-MNIST and CIFAR-10 datasets were used for training and evaluating the KAN-Mixers, MLP-Mixer, MLP and KAN models. Both the hyperparameter optimization and the training and evaluation of the models were performed using the k-fold cross-validation approach. Finally, the results were tabulated, considering the average accuracy.

\section{Results and discussion}

\subsection{Hyperparameter optimization}
The hyperparameter optimization of the KAN-Mixers model was performed using the random search method for the hyperparameters patch size, dim size, layers and learning rate, with the aim of maximizing the accuracy value in validation data. To obtain the model performance, the k-fold cross-validation approach was used, with k equal to 5. In addition, the number of trials performed by the random search was set to 10, in other words, the random search randomly selected 10 different hyperparameter configurations for the KAN-Mixers. The possible values for each hyperparameter are specified in Table 1.

\begin{table}[!ht]
    \renewcommand{\tabcolsep}{18 pt}
    \centering
    \begin{tabular}{c c}
        \hline
        Hyperparameter & Values \\
        \hline
        Patch size & \{(4,4), (8, 8), (16,16)\} \\
        Dim size & \{64, 128, 256\} \\
        Layers & \{6, 8, 10\} \\
        Learning rate & between 0.0001 e 0.001 \\
        \hline
    \end{tabular}
    \caption{Values for each hyperparameter.}
\end{table}

As a result of the hyperparameter optimization, the best values were obtained for each hyperparameter. The best hyperparameter configuration obtained an average accuracy of 0.66. Therefore, the hyperparameter values contained in this configuration were used in the KAN-Mixers model and are described in Table 2.

\begin{table}[!ht]
    \renewcommand{\tabcolsep}{18 pt}
    \centering
    \begin{tabular}{c c}
        \hline
        Hyperparameter & Values \\
        \hline
        Patch size & (4,4) \\
        Dim size & 256 \\
        Layers & 8 \\
        Learning rate & 0.00012820100418916918 \\
        \hline
    \end{tabular}
    \caption{Best hyperparameters.}
\end{table}

\subsection{Experiment on Fashion-MNIST dataset}
All models were trained using the k-fold cross-validation approach for 50 epochs, with a batch size of 64, the Adam optimizer, and the CrossEntropyLoss loss function. The KAN-Mixers hyperparameters were defined using hyperparameter optimization. The MLP-Mixer model had the same hyperparameters as the KAN-Mixers model, except for the learning rate, which was set to the default value of 0.001. The MLP and KAN models kept all their default hyperparameters. In addition, the RandomHorizontalFlip and RandomRotation data augmentations were applied. Finally, the resolution of the input images was standardized to 32x32. The results of training all models on the Fashion-MNIST dataset, in terms of accuracy, can be seen in the Violin and Strip plots in Figure 5.

\begin{figure}[!ht]
  \centering
  \includegraphics[width=0.70\textwidth]{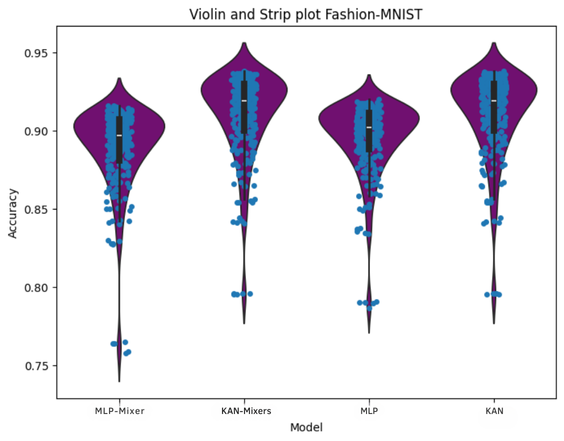}
  \caption{Violin and Strip plots of model training on the Fashion-MNIST set.}
  \label{fig:fig5}
\end{figure}

Observing the graphs in Figure 5, it can be seen that the KAN-Mixers and KAN models present a high density of accuracy values between 0.90 and 0.95, with some outliers between 0.85 and 0.80. The MLP-Mixer and MLP models showed a high concentration of accuracy values close to 0.90. It can also be seen that the models based on KAN layers presented a median (white line) higher than the median of the models based on MLP, which may indicate a slight superiority of the KAN layers in relation to the MLP layers.

\subsubsection{Accuracy and statistical significance testing}
The overall performance of machine learning models for the classification task can be measured by accuracy, which represents the level of accuracy of each model in a data set. Using validation data and a 5-fold cross-validation strategy, the average accuracy of the models and the Wilcoxon statistical significance test, for \textit{p}-values equal to 0.05 and 0.10, are presented in Table 3. In the statistical significance columns, the symbols =, + and - represent that a model is statistically equivalent, inferior or superior to the KAN-Mixers model, respectively. In the last column of Table 3, we present the percentage difference of each model compared to the KAN-Mixers model.

\begin{table}[!ht]
    \renewcommand{\tabcolsep}{5 pt}
    \centering
    \begin{tabular}{c c c c c c}
        \hline
        Model & Average accuracy & Standard deviation & \textit{p} = 0.05 & \textit{p} = 0.10 & Difference (\%)\\
        \hline
        MLP & 0.8873 & 0.0047 & = & + & 1.74\%\\
        KAN & 0.8916 & 0.0041 & = & + & 1.26\%\\
        MLP-Mixer & 0.8980 & 0.0015 & = & + & 0.55\%\\
        KAN-Mixers & \textbf{0.9030} & 0.0033 & & & \\
        \hline
    \end{tabular}
    \caption{Average accuracy of each model on the Fashion-MNIST dataset.}
\end{table}

As can be seen in Table 3, the KAN-Mixers model achieves an average accuracy performance slightly higher than that obtained by the MLP-Mixer model, while outperforming the MLP and KAN models in terms of absolute values. When verifying the statistical significance of the difference between the models, it was observed that the KAN-Mixers is equivalent to its competitors for a \textit{p}-value equal to 0.05. For a \textit{p}-value equal to 0.10, it is possible to conclude that the KAN-Mixers model is superior to its competitors in the Fashion-MNIST dataset.

\subsection{Experiment on CIFAR-10 dataset}
To perform the experiments on the CIFAR-10 dataset, the hyperparameters used in each model were the same as those used in the experiment on the Fashion-MNIST dataset. The training results using cross-validation for all models in the CIFAR-10 dataset, considering accuracy , can be seen in Figure 6.

\begin{figure}[!ht]
  \centering
  \includegraphics[width=0.70\textwidth]{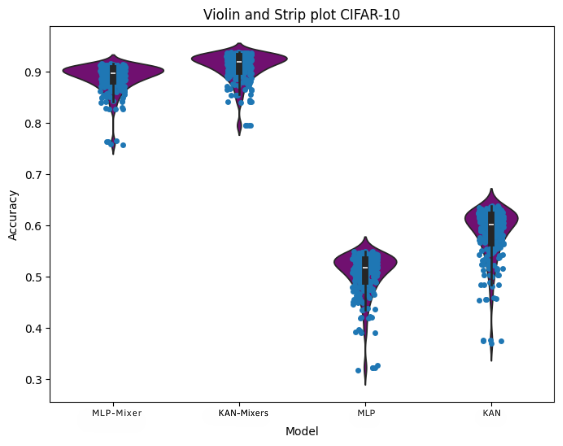}
  \caption{Violin and Strip plots of model training on the CIFAR-10 set.}
  \label{fig:fig6}
\end{figure}

The concentration of accuracy values obtained in the training of each model can be observed when analyzing Figure 6. Thus, it can be seen that the MLP-Mixer and KAN-Mixers models presented the majority of accuracy values obtained close to 0.90, with the KAN-Mixers model presenting a median higher than that obtained by the MLP-Mixer model. The KAN and MLP models presented the highest concentrations of accuracy values between 0.50 and 0.65, with the KAN model obtaining a median higher than the median of the MLP model.

\subsubsection{Accuracy and statistical significance testing}
Using the 5-fold cross-validation approach, the average accuracy of the models and the Wilcoxon statistical significance test, for \textit{p}-values equal to 0.05 and 0.10, are presented in Table 4. In the statistical significance columns, the symbols =, + and - represent that a model is statistically equivalent, inferior or superior to the KAN-Mixers model, respectively.

\begin{table}[!ht]
    \renewcommand{\tabcolsep}{5 pt}
    \centering
    \begin{tabular}{c c c c c c}
        \hline
        Model & Average accuracy & Standard deviation & \textit{p} = 0.05 & \textit{p} = 0.10 & Difference (\%)\\
        \hline
        MLP & 0.5055 & 0.0040 & = & + & 27.58\%\\
        KAN & 0.5400 & 0.0063 & = & + & 22.64\%\\
        MLP-Mixer & 0.6741 & 0.0054 & = & + & 3.42\%\\
        KAN-Mixers & \textbf{0.6980} & 0.0069 &  &  &\\
        \hline
    \end{tabular}
    \caption{Average accuracy of each model on the CIFAR-10 dataset.}
\end{table}

As shown in Table 4, the KAN-Mixers model outperformed the MLP-Mixer, MLP and KAN models, with a difference of 0.0239, 0.1925 and 0.1580, respectively. When assessing the statistical significance between the accuracies of the models, it was found that the KAN-Mixers model was considered equivalent to its competitors for a \textit{p}-value equal to 0.05. On the other hand, for a \textit{p}-value equal to 0.10, it is possible to state that the KAN-Mixers model is superior to its competitors.

Given that the CIFAR-10 dataset is more challenging than the Fashion-MNIST dataset, and that the KAN-Mixers and KAN models were superior to the MLP-Mixer and MLP models, considering the average accuracy, it is possible to note strong evidence that corroborates the research hypothesis presented in this work, that the KAN layers are more appropriate for feature extraction in images than the MLP layers, and are also in agreement with the results presented in the Fashion-MNIST dataset. Given this, the KAN-Mixers model can be an alternative to the MLP-Mixer, MLP and KAN models, especially when there are no restrictions regarding the time required to train the models, since the computational cost is a deficiency of the KAN-Mixers model.

\section{Conclusion}

In this work, a new deep learning model for image classification was developed, called KAN-Mixers. For this purpose, a literature review on MLP-Mixer and KANs was carried out, which allowed the extraction of the architectural details necessary for the development of the KAN-Mixers model. Furthermore, aiming to alleviate the main deficiency of KANs, which is the higher computational cost when compared to the MLP model, a more efficient implementation of the KAN network was used.

To perform the experiments, the Fashion-MNIST and CIFAR-10 datasets were used. We compared the KAN-Mixers, MLP-Mixer, MLP and KAN models using the Precision, Recall, F1-Score and Accuracy performance metrics. The results obtained showed that the KAN-Mixers model performed competitively when compared to the MLP-Mixer model on the Fashion-MNIST dataset. For the CIFAR-10 dataset, the proposed model obtained the best results when compared to its competitors, presenting an average accuracy of approximately 0.70. 

\section{Acknowledgments}
This study was financed in part by the Conselho Nacional de Desenvolvimento Científico e Tecnológico (CNPq), and in part by the Manna Team.

\bibliographystyle{unsrt}  
\bibliography{references}

\end{document}